\documentclass[10pt,twocolumn,letterpaper]{article}

\usepackage{cvpr}              
\usepackage{multirow}
\usepackage{graphicx}
\usepackage{tabularx}
\usepackage{array}
\usepackage{wrapfig}
\usepackage{colortbl}

\usepackage[numbers]{natbib}

\definecolor{cvprblue}{rgb}{0.21,0.49,0.74}
\usepackage[pagebackref,breaklinks,colorlinks,allcolors=cvprblue]{hyperref}
\newcommand{\ourmodel}{GeoFlow}

\def\confName{CVPR}
\def\confYear{2026}

\title{GeoFlow: Real-Time Fine-Grained Cross-View Geolocalization via Iterative Flow Prediction}

\author{
Ayesh Abu Lehyeh$^{1}$, Xiaohan Zhang$^{1}$, Ahmad Arrabi$^{1}$,
Waqas Sultani$^{2}$, Chen Chen$^{3}$, Safwan Wshah$^{1*}$ \\
$^{1}$ Vermont Artificial Intelligence Lab, Department of Computer Science, University of Vermont \\
$^{2}$ Intelligent Machines Lab, Information Technology University \\
$^{3}$ Institute of Artificial Intelligence, University of Central Florida \\
\small
$^{*}$ Corresponding and senior author.
}

\begin{document}
\maketitle
\begin{abstract}
Accurate and fast localization is vital for safe autonomous navigation in GPS-denied areas. Fine-Grained Cross-View Geolocalization (FG-CVG) aims to estimate the precise 2-Degree-of-Freedom (2-DoF) location of a ground image relative to a satellite image. However, current methods force a difficult trade-off, with high-accuracy models being slow for real-time use. In this paper, we introduce \ourmodel{}, a new approach that offers a lightweight and highly efficient framework that breaks this accuracy-speed trade-off. Our technique learns a direct probabilistic mapping, predicting the displacement (in distance and direction) required to correct any given location hypothesis. This is complemented by our novel inference algorithm, Iterative Refinement Sampling (IRS). Instead of trusting a single prediction, IRS refines a population of hypotheses, allowing them to iteratively `flow' from random starting points to a robust, converged consensus. Even its iterative nature, this approach offers flexible inference-time scaling, allowing a direct trade-off between performance and computation without any re-training. Experiments on the KITTI and VIGOR datasets show that \ourmodel{} achieves state-of-the-art efficiency, running at real-time speeds of~29 FPS while maintaining competitive localization accuracy. This work opens a new path for the development of practical real-time geolocalization systems. \textbf{Code is available at:} \href{https://github.com/AyeshAbuLehyeh/GeoFlow}{GitHub}
\end{abstract} 
\vspace{-12pt}
\section{Introduction}
Intelligent systems such as autonomous vehicles, robotics, and augmented reality require precise and real-time localization to operate safely in complex environments. 
While satellite-based positioning approaches (e.g., GPS or Galileo) are traditionally used, they proved to be unreliable in many practical settings where signals are blocked or affected by surrounding objects~\citep{survey, ubigtloc, sacp}. Recently, many attempts have been made for precise localization by leveraging Fine-Grained Cross-View Geolocalization (FG-CVG)~\citep{Fervers_2023_CVPR,lentsch2023slicematch,Shi_2023_ICCV,shi2022beyond,wang2023fine,fg2}. The general goal of Fine-Grained Cross-View Geolocalization (FG-CVG) is to find the full 3-DoF pose of a ground camera: its 2-DoF planar location $(x, y)$ and its 1-DoF orientation ($\theta$). However, in many real-world applications, such as a vehicle-mounted camera, the orientation is often known or can be reliably obtained from onboard sensors like an IMU or a compass. Therefore, in this work, we focus on the 2-DoF localization task. This problem has high practical relevance, as solving for the $(x, y)$ location is the foundational challenge for navigation in GPS-denied areas. Similar to many cross-view geo-localization tasks, FG-CVG is difficult due to extreme viewpoint variations, scale differences, and appearance differences between the views~\citep{survey,geodtr,zhang2024geodtr+,SAFA,Vigor}.

\begin{figure}[t]
    \vspace{-10pt}
    \centering
    \includegraphics[width=1.0\columnwidth]{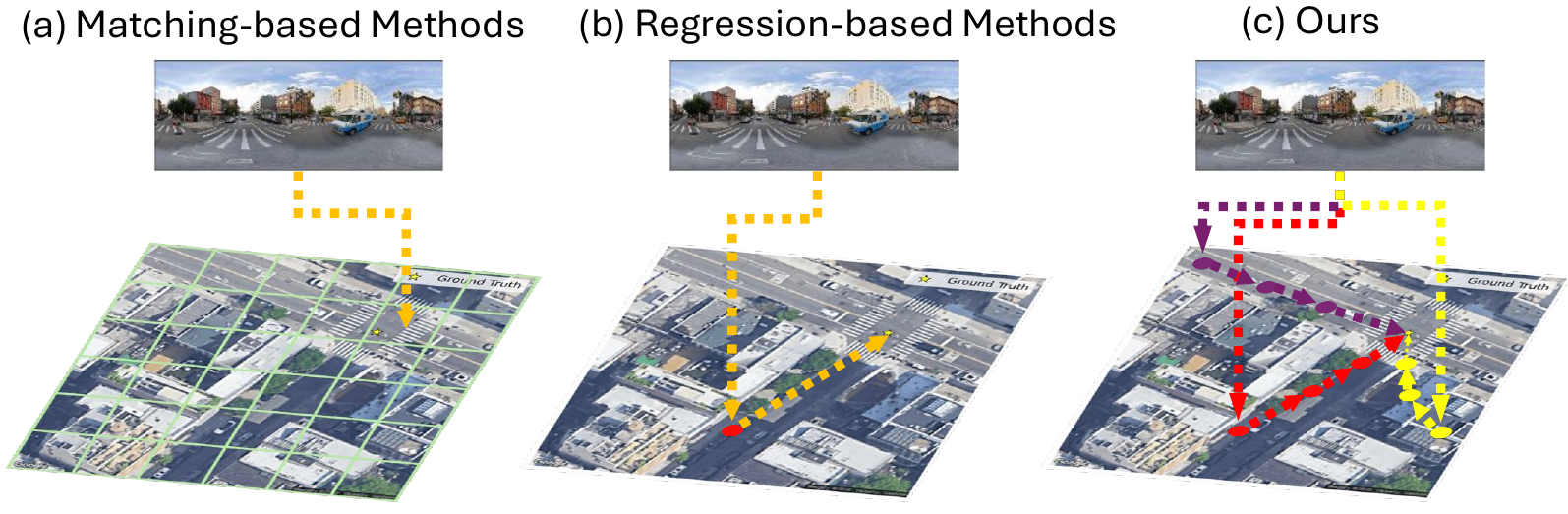} 
    \caption{Comparison of FG-CVG techniques. (a) Match-based methods search over discrete grids. (b) Regression-based methods directly or iteratively regress the pose. (c) Our method uses learned displacement predictions to iteratively guide multiple diverse hypotheses (in different colors) in continuous space towards the ground truth.}
    \label{fig:fgcvg_approach_comparison} 
    \vspace{-10pt}
\end{figure}

Existing FG-CVG methods can be categorized into two main types: matching-based methods~\citep{Fervers_2023_CVPR,lentsch2023slicematch,xia2022visual,matchingWACV,CCVPE, cbev} and regression-based methods~\citep{Shi_2023_ICCV,shi2022beyond,wang2023fine,Vigor,song2023learning}. Matching-based methods discretize the search space into a finite set of patches where a model is trained to predict a probability distribution for the ground image's location over these patches. However, this reliance on discretization inherently introduces quantization errors. Such errors can become more significant as the search area expands, thereby limiting localization accuracy and posing scalability challenges for larger satellite regions. While some of these methods optimize a Negative Log-Likelihood (NLL) loss to provide probabilistic uncertainty estimates, this valuable aspect is often compromised by the quantization artifacts inherent in their patch-based representations. 



Regression-based methods, on the other hand, operate within a continuous representation of the pose space. Simpler direct regression methods, for instance, may provide initial pose estimates but can struggle to achieve high fine-grained accuracy~\citep{Vigor}. More advanced methods aim for greater precision by iteratively refining an initial pose estimate through visual correspondence. However, deploying these methods can introduce practical hurdles. Some of these methods may require prior knowledge (such as accurate camera intrinsics)~\citep{shi2022beyond, song2023learning,fg2}, and others may require intermediate steps such as a bird's eye view (BEV) or homography estimation~\citep{song2023learning,fg2,wang2023fine}, increasing overall complexity and hindering real-time deployment. Furthermore, many of these methods usually produce only a deterministic point estimate of the pose, thereby lacking the uncertainty modeling needed to generate probabilistic estimates in continuous space~\citep{shi2022beyond}. Additionally, these methods typically lack flexible inference-time scaling, which is important for practical deployment. 

Considering these limitations, 
in this paper, we raise the following question: \textit{Is there a method that can solve FG-CVG in continuous space, achieve accurate localization, and operate at real-time speeds suitable for practical deployment?}



Flow matching models recently showed significant success in image synthesis~\citep{lipman2023flow,pmlr-v235-image}. Such models learn a vector field that transports samples from a prior distribution to a target distribution. In inference, a sample from a given prior distribution is updated iteratively by the predicted vector field until reaching the target distribution. In FG-CVG, this iterative refinement process mimics how humans solve localization problems. Humans do not typically jump to a precise location in a single step. Similarly, our approach starts with a rough coarse estimate and then iteratively refines based on visual cues (see~\Cref{fig:fgcvg_approach_comparison}).  

Inspired by this observation, in this paper, we propose a novel FG-CVG flow-inspired approach (\ourmodel{}) that performs regression from multiple initial pose (location) hypotheses toward the ground-truth location. Instead of learning a continuous flow field (like in flow matching), \ourmodel{} directly predicts the displacement (distance and direction) from each candidate pose to the target, effectively learning a regression field over the pose space. Additionally, to complement and enhance localization accuracy, we propose an Iterative Refinement Sampling (IRS) algorithm during inference. IRS generates diverse, random pose hypotheses and then iteratively refines them using the learned regression field. This process naturally guides the predictions towards more confident regions of the pose space, which also provides a practical measure of the overall localization confidence. By combining the proposed flow-inspired approach and IRS algorithm, \ourmodel{} achieves competitive geo-localization results on the popular VIGOR~\citep{Vigor} and KITTI~\citep{kitti} datasets. Moreover, by leveraging the flexibility of the IRS algorithm, we observe inference-time scaling behavior from \ourmodel{}, advancing the research of FG-CVG by introducing a novel paradigm for boosting the performance. While iterative methods are often computationally expensive, IRS is designed for high speed. We run the costly visual analysis \textit{only once}, and then apply an extremely fast iterative refinement which makes our IRS very efficient during inference.

\begin{figure*}[t!]
    \vspace{-15pt}
    \centering
    \includegraphics[width=2.0\columnwidth]{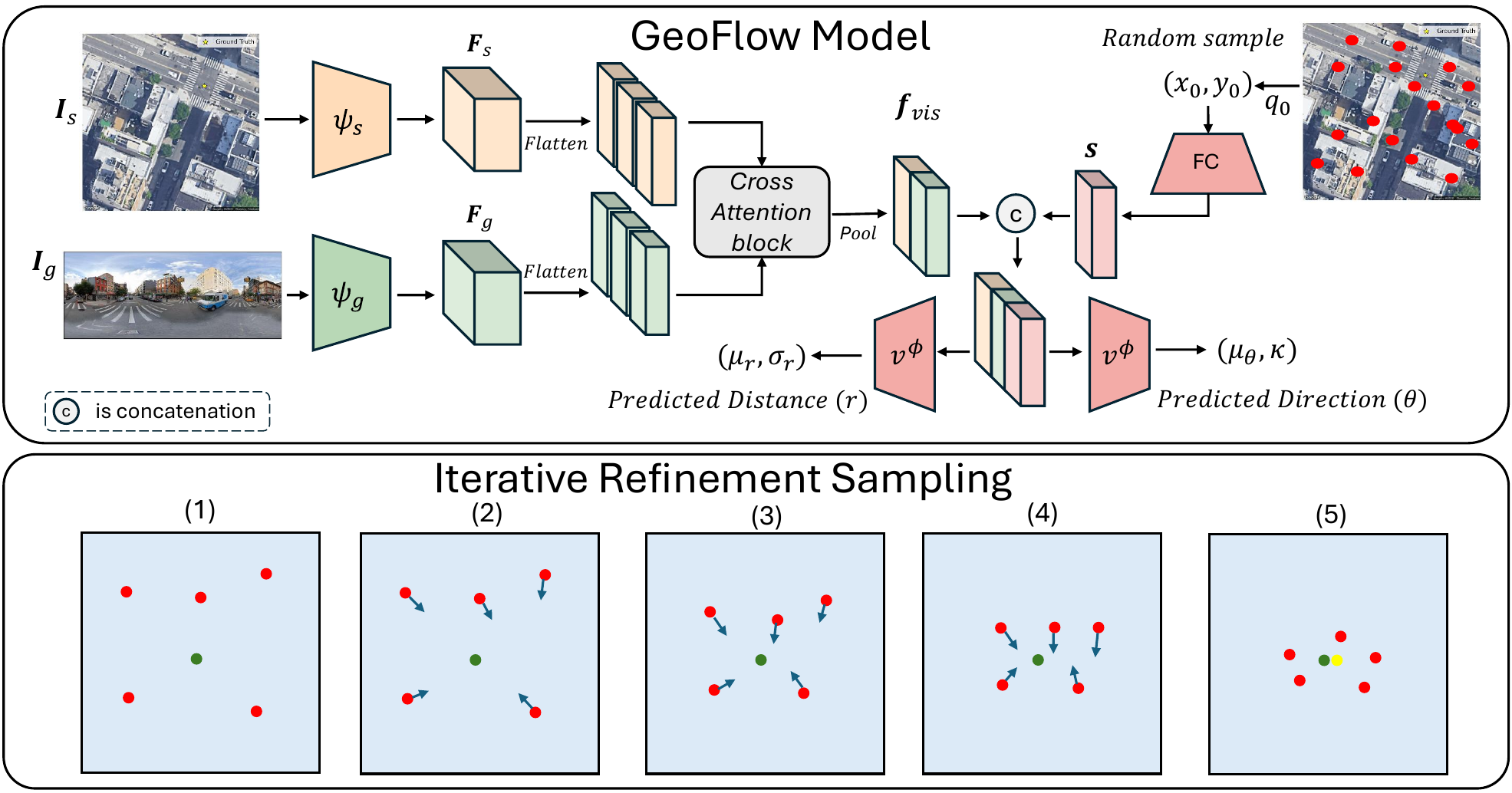} 
    \caption{The top panel illustrates the architecture of our proposed \ourmodel{}. Ground ($\mathbf{F}_g$) and satellite ($\mathbf{F}_s$) features are extracted by $\psi_g \text{ and } \psi_s$, then flattened, positionally encoded, and fused by a Cross-Attention block to produce a single visual representation, $\mathbf{f}_{vis}$. Separately, an initial location point $\mathbf{q}_0 = (x_0, y_0)$ is passed through a fully connected layer (FC). This point embedding is concatenated with $\mathbf{f}_{vis}$ to form the joint representation $\mathbf{s}$. Finally, two regression heads ($\mathbf{v}^\phi$) predict the parameters for the probabilistic displacement: distance $(\mu_r, \sigma_r)$ and direction $(\mu_\theta, \kappa)$. The bottom panel illustrates our Iterative Refinement Sampling (IRS) algorithm. (1) The process starts with $N$ randomly sampled hypotheses (red dots), the green dot represents the ground truth location. (2-4) In each round, our model predicts the displacement (blue arrows) for all hypotheses, which are then updated and iteratively converge toward the target. (5) After $R$ rounds, the final robust location (yellow dot) is computed as the mean of the entire converged population.}
    \label{fig:framework_overview}
    \vspace{-15pt}
\end{figure*}

In this work, our contributions can be summarized as threefold:
\begin{itemize}[leftmargin=*]
\setlength\itemsep{2pt}
    \item We propose \ourmodel{}, a novel flow-inspired regression approach designed for Fine-Grained Cross-View Geolocalization. \ourmodel{} operates directly in continuous pose space and is designed to predict a displacement vector (distance and direction) from multiple initial location hypotheses toward the ground-truth.
    \item We introduce a novel inference algorithm, IRS, which enhances localization accuracy and robustness by generating diverse pose hypotheses and iteratively refining them through the learned displacement fields. IRS not only improves convergence toward the true pose but also provides a natural estimate of localization confidence while maintaining low computational overhead.
    \item Through extensive experiments on VIGOR and KITTI datasets, we demonstrate that \ourmodel{} achieves competitive accuracy to SOTA methods while operating at a fraction of their computational cost. Notably, our model uses 7.8$\times$ fewer parameters and 4$\times$ GFLOPS reduction than SOTA methods like CCVPE~\citep{CCVPE}, making it ideal for real-time settings. Moreover, the IRS mechanism enables inference-time scalability, offering a controllable trade-off between accuracy and speed for real-time applications.
\end{itemize}

\section{Related Work}
\label{sec:related work}

\noindent\textbf{Cross-View Geolocalization (CVGL): } The foundational task in this domain is Cross-View Geolocalization, typically framed as an image retrieval problem~\citep{SAFA, CVUSA, liu2019lending, DSM, cvmnet, Vigor, l2ltr, geodtr, zhang2024geodtr+}. These methods learn global descriptors for ground and satellite images to find the best-matching satellite region for a ground-view query from a large database~\citep{shi2022beyond}. In these cross-view retrieval methods, the camera is assumed to be located at the center of the satellite map. However, this assumption limits the generalization of these methods and restricts their precision to the coarse level of the entire satellite tile.

\noindent\textbf{Fine-Grained Cross-View Geolocalization (FG-CVG): } 
Existing Fine-Grained Cross-View Geolocalization methods follow two main paradigms. Matching-based methods discretize the continuous pose space into a finite grid of patches~\citep{Fervers_2023_CVPR, xia2022visual, lentsch2023slicematch, cbev}, reframing the problem as a classification task. While this allows for probabilistic modeling, this "patchify" process has a fundamental limitation of quantization error. The model's precision is capped by the patch size, posing significant scalability challenges for real-world search areas. To overcome this, regression-based methods operate in a continuous space. Early work, such as VIGOR~\citep{Vigor}, introduced this concept by adding a regression head to a retrieval backbone but struggled to achieve high fine-grained accuracy. More advanced methods have emerged, but they often introduce significant computational overhead, hindering real-time deployment. For instance, Shi et al.~\citep{shi2022beyond} use a Levenberg-Marquardt (LM) optimization to iteratively refine the pose. However, this requires heavy geometric projections and relies on priors like camera intrinsics. Similarly, methods like \citep{fg2,song2023learning} leverage dense feature matching between the two views, but their geometric alignment modules also depend on these same heavy projections and intrinsic camera data. Other approaches, like~\citep{wang2023fine}, use recurrent homography estimation, which still requires an intermediate BEV projection. Finally, the work in~\citep{CCVPE} suffer from high memory utilization in their complex matching decoders, making them less appropriate for efficient, real-time deployment. In contrast to these approaches, we propose \ourmodel{}, a lightweight, flow-inspired model that regresses the camera pose in continuous space without any ''patchify'' process. Furthermore, our \ourmodel{} leverages the proposed IRS to fully explore the search area achieving robust consensus. Finally, for the first time, we observe inference time scaling on the FG-CVG task from our \ourmodel{}, enabling \ourmodel{} to achieve SOTA-comparable accuracy at real-time speeds, which implies its potential for practical deployment in autonomous systems.
\section{Methodology}
\label{sec:methods}

Figure~\ref{fig:framework_overview}  illustrates the proposed \ourmodel{}, designed for FG-CVG. Given a ground query image $\mathbf{I}_g \in \mathbb{R}^{H_g \times W_g \times 3}$ and a corresponding geo-referenced satellite map $\mathbf{I}_{s} \in \mathbb{R}^{H_{s} \times W_{s} \times 3}$, the objective of FG-CVG is to estimate the precise 2-DoF location $\mathbf{q} = (x, y)$ of the ground camera relative to the reference satellite map's coordinate system. To achieve this, \ourmodel{} is a flow-inspired model that performs regression from multiple initial pose hypotheses to perform pose estimation directly within the continuous satellite space. Specifically, \ourmodel{} learns a regression field $\mathbf{v}^\phi(\mathbf{q}_0,\mathbf{I}_g, \mathbf{I}_s)$ that predicts the probabilistic displacement vector (distance $r$ and direction $\theta$) from an arbitrary initial hypothesis $\mathbf{q}_0 \sim \mathbf{p}_0$ towards the target location $\mathbf{q}_1$. Critically, the model predicts parameters that define a probabilistic distribution over this displacement, allowing a Negative Log-Likelihood (NLL) training objective for both $r$ and $\theta$. Furthermore, \ourmodel{} captures localization confidence through the novel IRS algorithm during inference. This multi-seed sampling approach can iteratively refine pose estimates using the learned displacement vector and evaluate the consensus of the hypotheses, yielding a robust final location. The following subsections detail the cross-view feature extraction, the probabilistic displacement regression, the training objective, and the IRS technique.

\subsection{Cross-View Feature Extraction and Matching}
\label{subsec:feature_extraction_matching}

We utilize two separate backbones, defined as $\psi_g$ and $\psi_s$, to extract features from the ground image $\mathbf{I}_g$ and the satellite map $\mathbf{I}_s$, respectively. We employ the lightweight EfficientNet-B0~\citep{Tan2019EfficientNetRM} architecture as the backbone for both $\psi_g$ and $\psi_s$. This was an intentional choice to prioritize an efficient model design and to clearly demonstrate the effectiveness of our proposed \ourmodel{}.
Denoting the output from the backbones as $\mathbf{F}_g' = \psi_g(\mathbf{I}_g)$ and $\mathbf{F}_s' = \psi_s(\mathbf{I}_s)$, these feature maps are first projected to a common dimension $d$ using $1 \times 1$ convolutional layers, yielding $\mathbf{F}_g \in \mathbb{R}^{H'_g \times W'_g \times d}$ and $\mathbf{F}_s \in \mathbb{R}^{H'_s \times W'_s \times d}$.

\noindent To inject explicit spatial awareness, these feature maps are augmented with fixed 2D sinusoidal positional encodings following~\citep{matchingWACV}:
\begin{equation}
\tilde{\mathbf{F}}_g = \mathbf{F}_g + \mathbf{PE}_g, \quad \tilde{\mathbf{F}}_s = \mathbf{F}_s + \mathbf{PE}_s.
\label{eq:pos_encoding}
\end{equation}
To match features across views, we employ a cross-attention mechanism~\citep{matchingWACV}. The positionally-encoded features are first flattened into sequences of tokens, $\mathbf{T}_g \in \mathbb{R}^{N_g \times d}$ (where $N_g = H'_g \times W'_g$) and $\mathbf{T}_s \in \mathbb{R}^{N_s \times d}$ (where $N_s = H'_s \times W'_s$). 
The core idea is to use the ground tokens $\mathbf{T}_g$ as queries to retrieve information from the satellite tokens $\mathbf{T}_s$, which serve as keys and values. The Query, Key, and Value projections are created as follows:
\begin{equation}
\mathbf{Q}_g = \mathbf{T}_g \mathbf{W}^Q, \quad \mathbf{K}_s = \mathbf{T}_s \mathbf{W}^K, \quad \mathbf{V}_s = \mathbf{T}_s \mathbf{W}^V,
\label{eq:qkv}
\end{equation}
where $\mathbf{W}^Q$, $\mathbf{W}^K$, and $\mathbf{W}^V$ are learnable weight matrices. The attention scores are computed by taking the scaled dot-product of the queries and keys, which are then normalized using the Softmax function:
\begin{equation}
\text{Scores} = \text{softmax}\left(\frac{\mathbf{Q}_g \mathbf{K}_s^T}{\sqrt{d_k}}\right),
\label{eq:scores}
\end{equation}
where $d_k$ is the dimension of the keys. The resulting attended features, $\mathbf{f}_g$, are obtained by taking the weighted sum of the values:
\begin{equation}
\mathbf{f}_g = \text{Scores} \times \mathbf{V}_s.
\label{eq:weighted_sum}
\end{equation}
This operation effectively creates a new ground-view representation $\mathbf{f}_g$ where each token is enriched with relevant fused information from the satellite view.
Finally, we apply adaptive average pooling to aggregate this representation into a single, fixed-dimensional vector:
\begin{equation}
\mathbf{f}_{vis} = \text{AvgPool}({\mathbf{f}_g}) \in \mathbb{R}^{d}.
\label{eq:pooling}
\end{equation}
This global visual representation $\mathbf{f}_{vis}$ serves as the core visual context that conditions the probabilistic regression network, as detailed in the following section.

\subsection{Probabilistic Displacement Regression}
\label{subsec:displacement_regression}

We frame FG-CVG as learning a probabilistic regression field, $\mathbf{v}^\phi(\mathbf{q}_0, \mathbf{f}_{vis})$, that directly estimates the 2-DoF displacement from an arbitrary initial pose hypothesis $\mathbf{q}_0 = (x_0, y_0)$ towards the ground truth location $\mathbf{q}_1 = (x_1, y_1)$. 
Our model relies on two key inputs: the global visual representation $\mathbf{f}_{vis}$ (from Sec.~\ref{subsec:feature_extraction_matching}) and the spatial location randomly sampled $\mathbf{q}_0$. The 2D coordinate $\mathbf{q}_0$ is first passed through a coordinate projection layer to produce a high-dimensional positional embedding $\mathbf{s}$. This spatial embedding is then concatenated with the visual representation $\mathbf{f}_{vis}$ to form a single, joint vector:
\begin{equation}
\mathbf{z} = [\mathbf{f}_{vis} \oplus \mathbf{s}].
\label{eq:fusion}
\end{equation}
This joint representation $\mathbf{z}$, which encapsulates both the scene's appearance and the current query location, is then fed into the main regression network. This network is implemented as a Multi-Layer Perceptron (MLP) that decodes this information to predict a probabilistic displacement.
Specifically, the MLP branches into two separate prediction heads to define the displacement in polar coordinates $(r, \theta)$ and its associated uncertainty. The distance head predicts the parameters $(\mu_r, \sigma_r^2)$ for a Gaussian distribution $\mathcal{N}(\mu_r, \sigma_r^2)$ over the distance $r$. Here, $\mu_r$ represents the predicted mean distance, and $\sigma_r^2$ is the predicted variance. Concurrently, the direction head predicts parameters for the von Mises-Fisher (vMF) distribution~\citep{vonmises}. The vMF distribution is a distribution on the hypersphere, analogous to the Gaussian for linear data. In our case, for a 2D displacement, the direction vector lies on the unit circle $S^1$ ($n=2$ dimensions in the general vMF definition). The probability density for a direction vector $\mathbf{u} \in S^1$ is given by:
\begin{equation}
p(\mathbf{u} | \mathbf{\mu}_\theta, \kappa) = C_2(\kappa) \exp(\kappa \mathbf{\mu}_\theta^T \mathbf{u}),
\label{eq:vmf}
\end{equation}
where $\mathbf{\mu}_\theta \in S^1$ is the mean direction vector, $\kappa \ge 0$ is the concentration parameter, and $C_2(\kappa)$ is the normalization constant, which for $n=2$ simplifies to $C_2(\kappa) = (2\pi I_0(\kappa))^{-1}$, with $I_0$ being the modified Bessel function of the first kind at order 0. Our network head is trained to predict the components of the mean direction $\mathbf{\mu}_\theta$ and the concentration $\kappa$.

\noindent This probabilistic formulation, with dedicated heads for distance and direction, is essential for our NLL training objective (as detailed in Sec~\ref{subsec:training_objective}), which is designed to match the parameters of these predicted distributions to the ground truth displacement.

During inference, this learned regression field provides a single-step refinement operation. Given a hypothesis $\mathbf{q}_0$, the network predicts the mean displacement parameters $(\mu_r, \mathbf{\mu}_\theta)$. The refined pose $\hat{\mathbf{q}}_1$ is then computed by applying this displacement:
\begin{equation}
\hat{\mathbf{q}}_1 = \mathbf{q}_0 + \mu_r \cdot \frac{\mathbf{\mu}_\theta}{||\mathbf{\mu}_\theta||_2},
\label{eq:single_step_refinement}
\end{equation}
This single-step refinement forms the core operation of our Iterative Refinement Sampling (IRS) algorithm, which, as detailed in Sec~\ref{subsec:irs}, leverages this field to robustly evaluate multiple hypotheses and determine a final, confident location.

\subsection{Training Objective}
\label{subsec:training_objective}

To train our regression network, we employ a probabilistic objective that minimizes the NLL of the ground truth displacement. This approach encourages the predicted distributions (from Sec.~\ref{subsec:displacement_regression}) to precisely match the true displacement in both distance and direction.

For each training sample $(\mathbf{I}_g, \mathbf{I}_s, \mathbf{q}_{gt})$, we first sample a random hypothesis $\mathbf{q}_0$ uniformly from the satellite map. From this pair, we compute the ground truth displacement vector $\mathbf{u}_{gt} = \mathbf{q}_{gt} - \mathbf{q}_0$. This vector is then decomposed into its polar components, which serve as the targets for our two specialized loss functions: the ground truth distance $r_{gt} = ||\mathbf{u}_{gt}||_2$ and the ground truth direction $\mathbf{\theta}_{gt} = \mathbf{u}_{gt} / r_{gt}$.

First, for the distance, we use the standard Gaussian NLL loss, $\mathcal{L}_r$. This loss encourages the predicted mean $\mu_r$ to match $r_{gt}$, while the predicted variance $\sigma_r^2$ regularizes the loss based on confidence:
\begin{equation}
\mathcal{L}_r = \frac{1}{2} \left( \frac{(r_{gt} - \mu_r)^2}{\sigma_r^2} + \log \sigma_r^2 \right).
\label{eq:loss_r}
\end{equation}
The first term is the squared error weighted by the inverse variance, which penalizes errors more heavily when the model is confident (small $\sigma_r^2$). The second term, $\log \sigma_r^2$, is a regularizer that prevents the network from collapsing the variance to zero.

Second, for the direction, we use the Angular von Mises-Fisher (AngMF) loss from~\citep{vonmiseloss}. This NLL loss is designed to directly minimize the angular error, which is more robust than simple L2 losses. The loss $\mathcal{L}_{\theta}$ is formulated as a function of the predicted concentration $\kappa$ and the angular error between the predicted mean direction and the ground truth:
\begin{equation}
\mathcal{L}_{\theta} = -\log(\kappa^2 + 1) + \kappa \cdot \cos^{-1}(\mathbf{\mu}_{\theta}^T \cdot \mathbf{\theta}_{gt}) + \log(1 + \exp(-\kappa \pi)).
\label{eq:loss_theta}
\end{equation}
Here, $\mathbf{\theta}_{gt}$ is the ground truth direction vector, and the term $\cos^{-1}(\mathbf{\mu}_{\theta}^T \cdot \mathbf{\theta}_{gt})$ represents the angular error. Similar to the distance loss, the main term $\kappa \cdot \cos^{-1}(\cdot)$ penalizes angular errors more heavily when the model is confident (high $\kappa$), while the other terms act as regularizers on the concentration parameter.

Finally, the total training loss $\mathcal{L}$ is the sum of these two components. By minimizing this combined objective, the network learns to predict both the correct displacement and a meaningful measure of confidence for each part.
\begin{equation}
\mathcal{L} = \mathcal{L}_r + \mathcal{L}_\theta.
\label{eq:total_loss}
\end{equation}

\subsection{Iterative Refinement Sampling (IRS)}
\label{subsec:irs}

While the NLL training objective (Sec.~\ref{subsec:training_objective}) teaches our network $\mathbf{v}^\phi$ to predict a displacement, a single prediction from an arbitrary starting point can be sensitive to visual ambiguities. To overcome this and achieve a highly robust final estimate, we introduce our novel inference algorithm, Iterative Refinement Sampling (IRS). This is a flexible, multi-step procedure that not only enhances localization accuracy but also provides a direct trade-off between performance and computation speed at inference time.
Instead of trusting a single prediction, IRS operates on a population of hypotheses. The process begins by initializing a set $\mathcal{Q}_0$ containing $N$ candidate poses, $\{\mathbf{q}_0^{(i)}\}_{i=1}^N$, drawn from a broad prior distribution (e.g., sampled uniformly across the satellite map).
The algorithm then iteratively refines this entire set of poses over $R$ rounds. In each round (indexed by $k=1...R$), every candidate pose $\mathbf{q}_{k-1}^{(i)}$ from the previous set $\mathcal{Q}_{k-1}$ is fed back into our learned regression network $\mathbf{v}^\phi$ (along with the scene's visual context $\mathbf{f}_{vis}$). The network predicts the displacement from that specific pose, and we apply the single-step refinement defined in Eq.~\ref{eq:single_step_refinement} to compute the updated pose $\mathbf{q}_{k}^{(i)}$:
\begin{equation}
\mathbf{q}_{k}^{(i)} = \mathbf{q}_{k-1}^{(i)} + \mu_r^{(i)} \cdot \frac{\mathbf{\mu}_\theta^{(i)}}{||\mathbf{\mu}_\theta^{(i)}||_2},
\label{eq:irs_step}
\end{equation}
where $(\mu_r^{(i)}, \mathbf{\mu}_\theta^{(i)})$ are the mean parameters predicted by the network $\mathbf{v}^\phi$ when conditioned on the previous pose $\mathbf{q}_{k-1}^{(i)}$.
Afterward, this refinement step is applied to all $N$ candidates in parallel for $R$ iterations. This process allows the entire population of hypotheses to "flow" across the map and converge toward the most probable location(s) dictated by the learned regression field.
After the final round $R$, we are left with a set of $N$ refined poses, $\mathcal{Q}_R = \{\mathbf{q}_R^{(i)}\}_{i=1}^N$. The final, robust location estimate, $\hat{\mathbf{q}}_{\text{final}}$, is then simply computed as the mean of this entire final population. This multi-step, multi-hypothesis approach allows the model to find a strong consensus, effectively averaging out the noise from any single ambiguous prediction and improving localization accuracy.

\section{Experiments}
\label{sec:experiments}

\noindent\textbf{Datasets and Evaluation Metrics:}
Our experiments are conducted on two popular FG-CVG datasets: VIGOR and KITTI.
\begin{itemize}
    \item \textbf{VIGOR}~\citep{Vigor} provides $105,214$ ground-level panoramic images and $90,618$ aerial images collected across four U.S. cities. Each aerial image covers a $\sim 70\,\mathrm{m} \times 70\,\mathrm{m}$ ground region. We utilize the corrected 2-DoF ground truth camera pose labels from~\citep{lentsch2023slicematch}. Our evaluations are performed on both the same-Area and cross-Area settings.
    
    \item \textbf{KITTI}~\citep{kitti} contains forward-facing ground images with a limited field of view, augmented with aerial images by~\citep{Yujiao}. Each aerial patch covers a region of $\sim 100\,\mathrm{m} \times 100\,\mathrm{m}$. We follow the standard data split and evaluate on both the same-area (Test1) and cross-area (Test2) test sets. As in prior works~\citep{lentsch2023slicematch, CCVPE}, we assume ground images are located within the central $40\,\mathrm{m} \times 40\,\mathrm{m}$ area of the corresponding aerial map.
\end{itemize}
For both datasets, we report the standard metrics for 2-DoF localization: mean and median translation error (in meters). For KITTI, we further decompose this error into longitudinal and lateral components, following~\citep{CCVPE}. As a key contribution of our work is efficiency, we also report and compare the inference speed (FPS) for all methods. {\color{cvprblue}\textit{For more details on the implementation, please refer to the supplementary material.}}

\begin{table*}[t]
\vspace{-15pt}
\centering
\caption{\textbf{KITTI} test results comparison. We report localization error (Mean/Median, m), lateral/longitudinal recall (R@1m/R@5m, \%), and inference speed (FPS) for both Same-Area and Cross-Area splits. Best results are in \textbf{bold}.}
\label{tab:main_results_kitti}
\setlength{\tabcolsep}{1.5pt}
\begin{tabular}{lccccccccccccc}
\toprule
& \multicolumn{1}{c}{Efficiency} & \multicolumn{6}{c}{Same-area} & \multicolumn{6}{c}{Cross-area} \\
\cmidrule(lr){2-2} \cmidrule(lr){3-8} \cmidrule(lr){9-14}
& FPS↑ & \multicolumn{2}{c}{$\downarrow$Loc. (m)} & \multicolumn{2}{c}{$\uparrow$Lateral (\%)} & \multicolumn{2}{c}{$\uparrow$Long. (\%)} & \multicolumn{2}{c}{$\downarrow$Loc. (m)} & \multicolumn{2}{c}{$\uparrow$Lateral (\%)} & \multicolumn{2}{c}{$\uparrow$Long. (\%)} \\
\cmidrule(lr){3-4} \cmidrule(lr){5-6} \cmidrule(lr){7-8} 
\cmidrule(lr){9-10} \cmidrule(lr){11-12} \cmidrule(lr){13-14}
Method &  & Mean & Median & R@1m & R@5m & R@1m & R@5m & Mean & Median & R@1m & R@5m & R@1m & R@5m \\
\midrule
GGCVT~\cite{Shi_2023_ICCV}        & 4.17 & –    & –    & 76.44 & 98.89 & 23.54 & 62.18 & –    & –    & 57.72 & 91.16 & 14.15 & 45.00 \\
CCVPE~\cite{CCVPE}        & 24.00 & 1.22 & 0.62 & 97.35 & 99.71 & 77.13 & 97.16 & 9.16 & \textbf{3.33} & 44.06 & 90.23 & 23.08 & 64.31 \\
HC-Net~\cite{wang2023fine}       & 25.00 & 0.80 & 0.50 & 99.01 & 99.73 & \textbf{92.20} & \textbf{99.25} & 8.47 & 4.57 & 75.00 & 97.76 & \textbf{58.93} & \textbf{76.46} \\
DenseFlow~\cite{song2023learning} & 7.30 & 1.48 & \textbf{0.47} & 95.47 & 99.79 & 87.89 & 94.78 & 7.97 & 3.52 & 54.19 & 91.74 & 23.10 & 61.75 \\
FG$^2$~\cite{fg2}          & 4.20 & \textbf{0.75} & 0.52 & \textbf{99.73} & \textbf{100.00} & 86.99 & 98.75 & \textbf{7.45} & 4.03 & \textbf{89.46} & \textbf{99.80} & 12.42 & 55.73 \\
\midrule
GeoFlow (Ours) & \textbf{29.49} & 0.98 & 0.68 & 96.85 & 99.68 & 74.05 & 98.75 & 8.42 & 5.60 & 36.36 & 83.85 & 14.76 & 52.51 \\
\bottomrule
\end{tabular}
\end{table*}


\subsection{Main Results}
\label{subsec:main_results}

\noindent\textbf{Quantitative Comparison.}
We compare \ourmodel{} against recent SOTA methods including GGCVT~\cite{Shi_2023_ICCV}, CCVPE~\cite{CCVPE}, HC-Net~\cite{wang2023fine}, DenseFlow~\cite{song2023learning}, and FG$^2$~\cite{fg2}. 

Our performance on the KITTI dataset, shown in \Cref{tab:main_results_kitti}, highlights our model's exceptional efficiency.
\ourmodel{} sets a new benchmark for speed, achieving 29.49 FPS. This is 1.2x faster than the next-fastest competitor (HC-Net) and 8-14x faster than compute-heavy models like FG$^2$.
In the same-area setting, our model proves that speed does not sacrifice accuracy. We remain highly competitive with the top performers, achieving a 0.98m mean error and 99.68\% lateral recall at 5m. This demonstrates an ideal balance of SOTA speed and robust, high-end performance.
In the challenging cross-area setting, our lightweight design remains remarkably robust. Our model's mean error is highly competitive, proving comparable to the fastest SOTA method, HC-Net, while achieving an 8.1\% error reduction over CCVPE. This result solidifies \ourmodel{} as a well-balanced solution, delivering comparable accuracy while being the most efficient model in the field. This performance trade-off is an expected result of our lightweight design.

On VIGOR (\Cref{tab:main_results_vigor_stacked}), \ourmodel{}'s efficiency stands out clearly. At 29.49 FPS, our model is~1.5x faster than the next-fastest methods (HC-Net and CCVPE) and~5-10x faster than highly accurate, compute-heavy models like FG$^2$. 
While heavyweight models like FG$^2$ lead in accuracy, our lightweight model remains highly competitive. We achieve a 2.5\% reduction in mean error over CCVPE in the same-area setting. More notably, in the challenging cross-area setting, our model's mean error outperforms CCVPE by 7.0\%, while also beating GGCVT and DenseFlow. This result confirms our model's excellent balance, delivering state-of-the-art speed while maintaining competitive localization accuracy. {\color{cvprblue}\textit{For more details on GeoFlow's 3-DoF extension, please refer to the supplementary material.}}

\begin{figure*}[t]
\vspace{-5pt}
    \centering
    \begin{minipage}[b]{0.95\textwidth}
        \centering
        \includegraphics[width=\columnwidth]{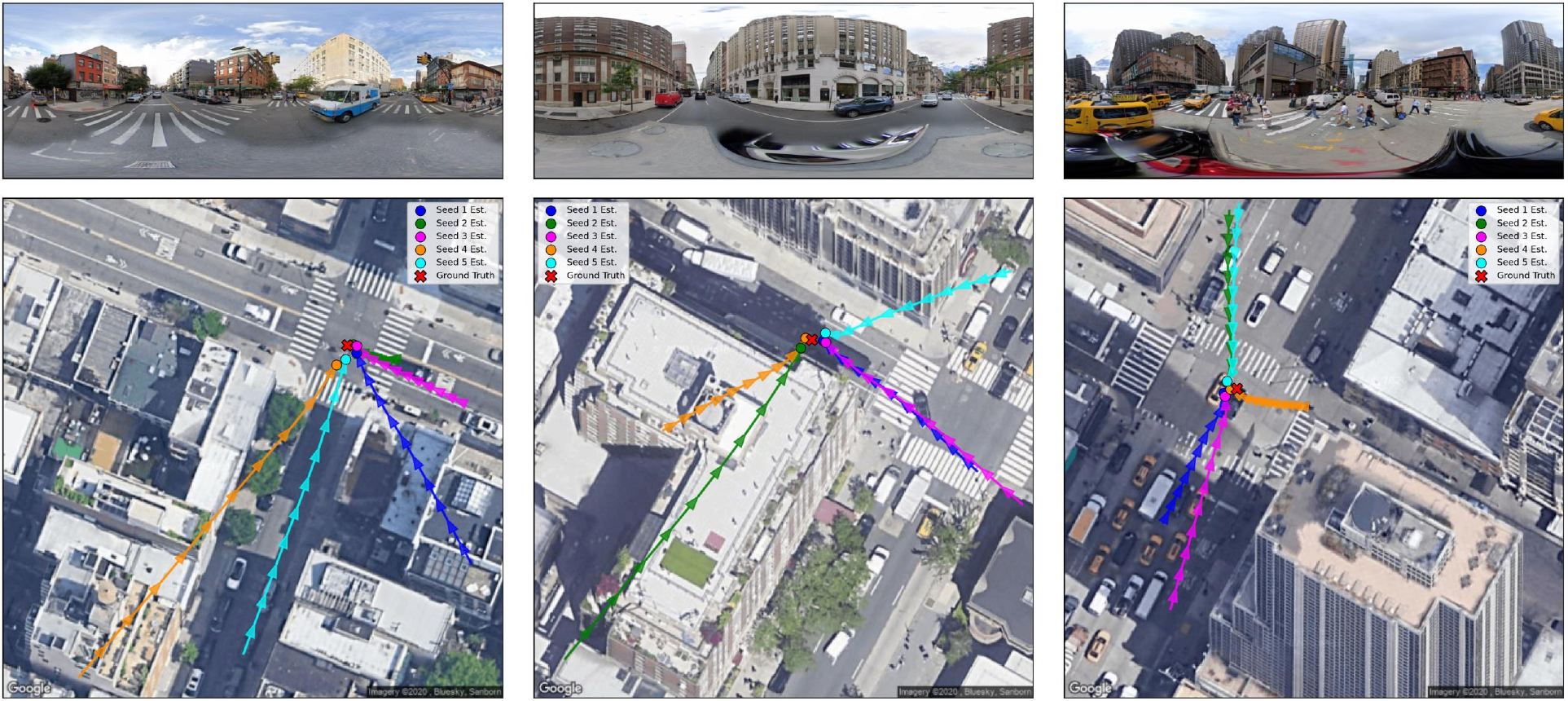}
        \caption{Qualitative examples on the \textbf{VIGOR} dataset. The trajectories demonstrate our model's regression field, showing how five randomly sampled initial hypotheses are guided towards a coherent solution near the ground truth (red X) using our IRS algorithm.}
        \label{fig:qualitative_results}
    \end{minipage}
    \hfill
    \begin{minipage}[b]{0.95\textwidth}
        \centering
        \includegraphics[width=\columnwidth]{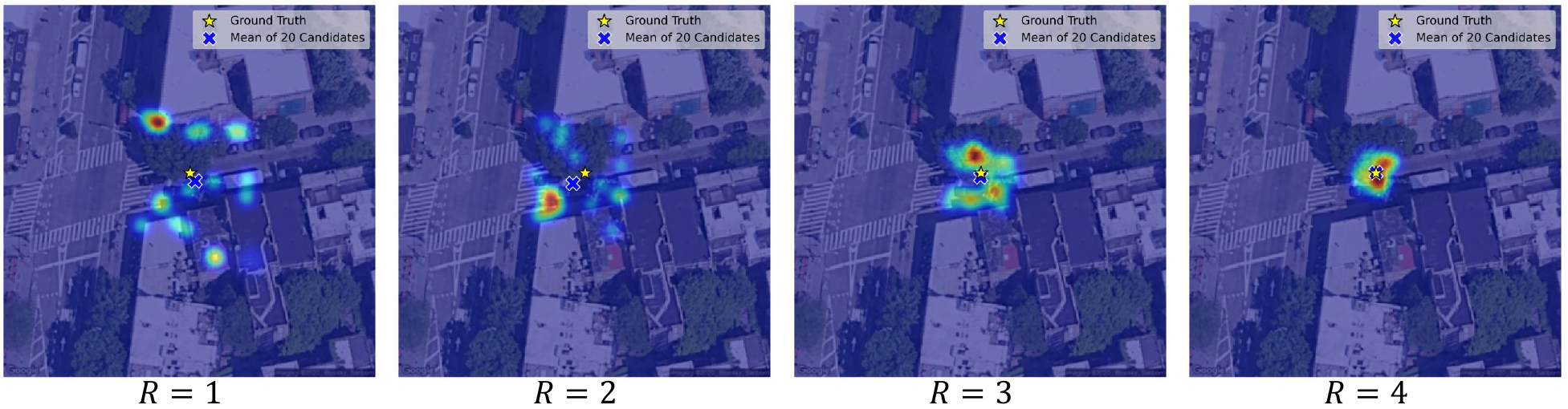}
        \caption{Visualization of an initial, scattered distribution of $N=10$ hypotheses is iteratively guided by our learned field over $R=4$ rounds in the IRS algorithm. The population progressively converges into a tight cluster centered on the Ground Truth.}
        \label{fig:irs_heatmap}
    \end{minipage}
    \vspace{-10pt}
\end{figure*}

\begin{table}[t]
\centering
\caption{Comparison on \textbf{VIGOR} dataset. Localization error (m) and inference speed (FPS) are reported. Best results are in \textbf{bold}.}
\label{tab:main_results_vigor_stacked}
\setlength{\tabcolsep}{1.5pt} 
\begin{tabular}{@{}l l c c c@{}} 
\toprule
Method & Area & $\downarrow$Mean (m) & $\downarrow$Med. (m) & $\uparrow$ FPS \\
\midrule
\multirow{2}{*}{GGCVT~\citep{Shi_2023_ICCV}} 
& Same & 4.12 & 1.34 & \multirow{2}{*}{3.57} \\
& Cross & 5.16 &  \textbf{1.40} & \\
\cmidrule(l){2-5}
\multirow{2}{*}{CCVPE~\citep{CCVPE}} 
& Same & 3.60 & 1.36 & \multirow{2}{*}{18.00} \\
& Cross &  4.97 & 1.68 & \\
\cmidrule(l){2-5}
\multirow{2}{*}{HC-Net~\citep{wang2023fine}} 
& Same & 2.65 & 1.17 & \multirow{2}{*}{20.00} \\
& Cross & 3.35 & 1.59 & \\
\cmidrule(l){2-5}
\multirow{2}{*}{DenseFlow~\citep{song2023learning}} 
& Same & 3.03 & \textbf{0.97} & \multirow{2}{*}{6.10} \\
& Cross & 5.01 & 2.42 & \\
\cmidrule(l){2-5}
\multirow{2}{*}{FG$^2$~\cite{fg2}} 
& Same & \textbf{2.18} & 1.18 & \multirow{2}{*}{3.60} \\
& Cross & \textbf{2.74} & 1.52 & \\
\cmidrule(l){2-5}
\multirow{2}{*}{\ourmodel{} (Ours)} 
& Same & 3.51 & 2.53 & \multirow{2}{*}{\textbf{29.49}} \\
& Cross & 4.62 & 2.78 & \\
\bottomrule
\end{tabular}
\end{table}

\noindent\textbf{Qualitative Results.}
In~\Cref{fig:qualitative_results}, we visualize the power of our learned regression field on VIGOR dataset. We show how five distinct, randomly sampled initial poses are refined using our IRS with ($R=10$). The trajectories demonstrate our model's robust ability to guide diverse, initial estimates towards a coherent solution near the ground truth.

In~\Cref{fig:irs_heatmap}, we visualize the confidence of our IRS algorithm. While our NLL loss provides a per-step uncertainty, IRS builds upon this to refine the final predictive confidence. The visualization shows how an initial, scattered distribution of $N$ hypotheses (at R=1) is refined iteratively. As the refinement rounds $R$ increase, the population of poses is actively guided by our learned regression field. This leads to a progressive "shrinking" and convergence of the hypotheses, transforming the broad initial set into a tight, focused cluster centered on the ground truth. This process illustrates how IRS builds confidence and achieves a robust consensus, actively sculpting the predictive uncertainty from a wide distribution into a sharp, final estimate. {\color{cvprblue}\textit{For more qualitative results, please refer to the supplementary material.}}

\begin{table}[h]
\vspace{-5pt}
    \centering
    \caption{Impact of refinement rounds ($R$) on KITTI Cross-Area.}
    \label{tab:ablation_R}
    \setlength{\tabcolsep}{3pt}
    \begin{tabular}{@{}cccc@{}}
    \toprule
    Rounds ($R$) & $\downarrow$Mean (m) & $\downarrow$Median (m) & $\uparrow$Speed (FPS) \\
    \midrule
    1 & 10.69 & 9.95 & \textbf{32.55} \\
    3 & 8.47 & 5.88 & 31.23 \\
    5 & 8.42 & 5.60 & 29.49 \\
    10 & \textbf{8.41} & \textbf{5.59} & 26.23 \\
    \bottomrule
    \end{tabular}
    \vspace{-5pt}
\end{table}

\begin{table}[h]
\vspace{-5pt}
    \centering
    \caption{Impact of initial seeds ($N$) on KITTI Cross-Area.}
    \label{tab:ablation_N}
    \setlength{\tabcolsep}{3pt}
    \begin{tabular}{@{}cccc@{}}
    \toprule
    Seeds ($N$) & $\downarrow$Mean (m) & $\downarrow$Median (m) & $\uparrow$Speed (FPS) \\
    \midrule
    1 & 8.58 & 5.75 & \textbf{30.70} \\
    5 & 8.49 & 5.66 & 30.05 \\
    10 & 8.42 & 5.60 & 29.49 \\
    20 & \textbf{8.41} & \textbf{5.60} & 28.08 \\
    \bottomrule
    \end{tabular}
    \vspace{-10pt}
\end{table}

\subsection{Inference-Time Scaling}
\label{subsec:inference_scaling}
A key advantage of our IRS algorithm is the ability to scale performance at inference-time \textit{without} any re-training. We can tune the number of seeds ($N$) and rounds ($R$) for a direct trade-off between accuracy and speed.
The magic of this approach lies in its efficiency. We run the EfficientNet backbone and the cross-attention module \textit{only once} to get the visual context $\mathbf{f}_{vis}$. The IRS iterations only re-run the tiny, lightweight coordinate projection layer and regression MLPs. This means we can add more initial seeds and refinement steps with minimal impact on our real-time speed.
We analyze this trade-off on the KITTI Cross-Area set in \Cref{tab:ablation_R} and \Cref{tab:ablation_N}. First, in \Cref{tab:ablation_R}, we fix $N=10$ seeds and vary the refinement rounds $R$. The results show a dramatic improvement when moving from a single step ($R=1$) to multi-step refinement. The mean error drops by 20.8\% at $R=3$. After this point, further refinement continues to offer smaller gains, improving the mean error to $8.41m$ at $R=10$. Similarly, in \Cref{tab:ablation_N}, we fix $R=5$ and vary the number of initial seeds $N$. We see a consistent, stable improvement as we increase $N$, with the mean error dropping from $8.58m$ ($N=1$) to $8.41m$ ($N=20$) as the consensus from the population of hypotheses grows stronger. In summary, this analysis shows our \ourmodel{}'s capability for inference-time scaling. We can achieve significant accuracy improvements by increasing $R$ and $N$ at test time, without any re-training. Notably, our lightweight model keeps the inference speed high (over 26 FPS) in all configurations. Based on these results, we select $N=10, R=5$ as our default, as it captures the vast majority of the accuracy gains while maintaining a high speed of 29.5 FPS.

\begin{table}[h]
    \centering
    \caption{Comparison of computational efficiency.}
    \label{tab:ablation_efficiency}
    \setlength{\tabcolsep}{1.5pt}
    \begin{tabular}{@{}lcccc@{}}
    \toprule
    Method & $\downarrow$Params(M) & $\downarrow$Mem(MiB) & $\uparrow$FPS & $\downarrow$Mean(m) \\
    \midrule
    CCVPE~\citep{CCVPE} & 57.40 & 4730 & 24.00 & 9.16 \\
    HC-Net~\citep{wang2023fine} & 11.21 & 1900 & 25.00 & 8.47 \\
    \midrule
    Ours & \textbf{7.38} & \textbf{686} & \textbf{29.49} & \textbf{8.42} \\
    \bottomrule
    \end{tabular}
    \vspace{-5pt}
\end{table}

\subsection{Computational Efficiency Analysis}
\label{subsec:computational_efficiency}

We provide a head-to-head comparison of computational efficiency on the KITTI dataset in \Cref{tab:ablation_efficiency}. All benchmarks were run on an NVIDIA V100 GPU. The results clearly demonstrate our model's superior efficiency.
In terms of model size, \ourmodel{} is 1.5x smaller than HC-Net and 7.8x smaller than CCVPE. This lightweight design leads to a significantly reduced memory footprint, using only 686 MiB, which is 2.7x less memory than HC-Net and 6.9x less than CCVPE.
Crucially, this efficiency does not come at the cost of performance. \ourmodel{} is the fastest method at 29.49 FPS, while also achieving the lowest mean error of $8.42m$ in this comparison. In summary, \ourmodel{} achieves a state-of-the-art balance of accuracy and efficiency, using a fraction of the parameters and memory while delivering top-tier speed and mean localization performance. {\color{cvprblue}\textit{For more details, please refer to the supplementary material.}}

\begin{table}[h]
    \centering
    \caption{Efficacy of the proposed IRS algorithm.}
    \label{tab:ablation_irs_efficacy}
    \setlength{\tabcolsep}{4pt}
    \begin{tabular}{@{}lcc@{}}
    \toprule
    IRS Config. & $\downarrow$Mean (m) & $\downarrow$Med (m) \\
    \midrule
    Single-Pass ($N=1, R=1$) & 12.47 & 11.79 \\
    Full IRS (Default) & \textbf{8.42} & \textbf{5.60} \\
    \bottomrule
    \end{tabular}
    \vspace{-5pt}
\end{table}

\subsection{Ablation Study}
\label{subsec:ablation_study}

Finally, to quantify the benefit of our inference strategy, we compare our full IRS algorithm against a simple single-pass baseline ($N=1, R=1$) on the KITTI dataset. The results in \Cref{tab:ablation_irs_efficacy} are definitive. Our full IRS process, with its robust multi-hypothesis consensus, dramatically improves localization, reducing the mean error by 32.5\% and more than halving the median error with a 52.5\% improvement. This confirms that our IRS algorithm is not a minor tweak but a critical component for achieving robust and accurate localization. {\color{cvprblue}\textit{For more ablation studies, please refer to the supplementary material.}}

\section{Conclusion and Future Work}

This paper introduced \ourmodel{}, a lightweight and highly efficient framework for fine-grained cross-view geolocalization. \ourmodel{} learns to predict the displacement (in distance and direction) required to correct any given location hypothesis. It is trained using a dedicated NLL loss for both the distance and direction, framing this as a probabilistic regression problem. Its name is inspired by its Iterative Refinement Sampling (IRS) algorithm, where a population of hypotheses iteratively \textit{flows} from random starting points to a robust, converged estimate, enabling high localization accuracy and flexible inference-time scaling without re-training. Despite its iterative nature, our experiments demonstrated that \ourmodel{} achieves a new state-of-the-art in computational efficiency, operating significantly faster than competing methods while maintaining competitive localization accuracy. This work's success opens new avenues for efficient, real-time geolocalization. Future efforts will focus on extending \ourmodel{} to full 6-DoF estimation and adapting its efficient principles for downstream tasks, like precise cross-view object geolocalization, further broadening its impact on real-world spatial understanding.

\section*{Acknowledgments}
This research has been supported by the U.S. National Science Foundation with grant 2119485.

{
    \small
    \bibliographystyle{ieeenat_fullname}
    \bibliography{main}
}

\maketitlesupplementary
\setcounter{section}{0}
\renewcommand{\thesection}{\Alph{section}}


In this supplementary material, we provide additional technical and experimental details, organized as follows:

\begin{itemize}
    \item \textbf{Implementation Details} ($\text{\cref{apdx:implemntation}}$). We offer more details about our final model architecture, training routines, hyperparameters and IRS parameters

    \item \textbf{Feasibility of 3-DoF Extension} ($\text{\cref{apdx:3dof_extension}}$). We present the methodology and quantitative results demonstrating GeoFlow's capability to generalize to the full 3-DoF pose estimation task with minimal overhead.

    \item \textbf{Backbone Capacity and Efficiency Trade-Off} ($\text{\cref{apdx:backbone_ablation}}$). We quantify the trade-off between model size and accuracy by comparing two EfficientNet backbones, justifying our final design choice for real-time deployment.
    
    \item \textbf{Efficiency Analysis and Deployment Considerations} ($\text{\cref{apdx:efficiency}}$). We analyze GeoFlow’s computational efficiency, including its decoupled inference design, unified efficiency benchmark, and implications for real-world edge deployment.
    
    \item \textbf{Uncertainty and Convergence Dynamics} ($\text{\cref{apdx:uncertain}}$). We offer visualizations showing how the IRS algorithm refines predictive uncertainty and builds confidence.    
    
    \item \textbf{Additional Qualitative Results} ($\text{\cref{apdx:vis}}$). We provide extensive visual evidence of the learned regression field, showcasing refinement trajectories in both same-area and cross-area settings.
    
    \item \textbf{Discussion of Failure Cases} ($\text{\cref{failure cases}}$). We provide an analysis of specific failure modes (e.g., highly vegetated areas) that contribute to the shift in the error distribution.
    
    \item \textbf{Societal Impact and Limitations} ($\text{\cref{apdx:impacts}}$). We provide the required discussion on the broader context, efficiency advantages, and practical limitations of our system.
\end{itemize}

\section{Implementation and Training Details}
\label{apdx:implemntation}

\noindent\textbf{Implementation Details:}
\ourmodel{} is implemented in PyTorch~\citep{paszke2019pytorch}. For both $\psi_g$ and $\psi_s$, we use an EfficientNet-B0~\citep{Tan2019EfficientNetRM} backbone, initialized with pre-trained ImageNet weights. 
The cross-attention module (Sec. 3.1 of the manuscript) is a standard multi-head attention layer with 4 heads. The coordinate projection layer embeds the 2D hypothesis into a 16-dimensional vector. This is concatenated with the 128-dimensional visual representation $\mathbf{f}_{vis}$, creating a 144-dimensional input for the regression MLPs. 

\noindent\textbf{Training Details:}
Our model is trained end-to-end for 200 epochs by optimizing the combined NLL loss (Eq. 11) on a single NVIDIA H100 GPU. We employ the AdamW optimizer~\citep{loshchilov2017decoupled} with a batch size of 80 and a weight decay of $1 \times 10^{-2}$. We use a differential learning rate: the pre-trained EfficientNet backbones are fine-tuned with a low learning rate of $1 \times 10^{-4}$, while the rest of the network is trained with a higher learning rate of $1 \times 10^{-3}$. For both VIGOR and KITTI, we resize the ground image $\mathbf{I}_g$ to $256 \times 1024$ and the aerial image $\mathbf{I}_s$ to $512 \times 512$. During training, the initial hypothesis $\mathbf{q}_0$ is formed by uniformly sampling a 2D translation $(x_0, y_0)$ from the valid spatial range. To ensure consistency across the model, all spatial inputs and outputs, including the initial hypothesis $\mathbf{q}_0$, the target location $\mathbf{q}_1$, are normalized to the continuous range $[-1, 1]$ relative to the reference satellite map. During inference, all reported inference results and comparisons (accuracy and FPS) are measured on a single NVIDIA V100 GPU to ensure a fair comparison with existing literature.

\noindent\textbf{Inference and IRS Parameters:}
During inference, we use our Iterative Refinement Sampling (IRS) algorithm (Sec.~3.4 in the main paper) to ensure a robust prediction. Unless otherwise stated, we adopt the following default parameters for evaluation: we initialize $N=10$ candidate poses (seeds) by sampling uniformly from the pose space. We then perform $R=5$ iterative refinement rounds, where each of the $N$ poses using Eq.~12.
The final location estimate is computed as the mean of all $N=10$ refined poses after the final round.


\begin{table*}[h]
\centering
\caption{3-DoF Pose Estimation Comparison on KITTI. GeoFlow demonstrates competitive accuracy against state-of-the-art methods while maintaining a superior computational profile. Best results are in \textbf{bold}}
\label{tab:3dof_results}
\setlength{\tabcolsep}{4pt}
\begin{tabular}{@{}llccccccc@{}}
\toprule
 & & \multicolumn{2}{c}{\textbf{Loc. (m)}} & \multicolumn{4}{c}{\textbf{Orientation ($\gamma$) Errors ($^{\circ}$)}} & \multicolumn{1}{c}{\textbf{Inference}} \\
\cmidrule(lr){3-4}\cmidrule(lr){5-8}\cmidrule(lr){9-9}
Dataset & Method & $\downarrow$Mean & $\downarrow$Median & $\downarrow$Mean & $\downarrow$Median & $\uparrow$R@1 & $\uparrow$R@5 & Speed (FPS) \\
\midrule
\multirow{3}{*}{\textbf{Same Area}} 
& FG$^2$~\cite{fg2} & \textbf{0.75} & \textbf{0.52} & \textbf{1.28} & \textbf{0.74} & \textbf{61.17}\% & \textbf{95.65}\% & 4.20 \\
& GeoFlow (3-DoF) & 1.03 & 0.69 & 2.51 & 1.93 & 27.30\% & 87.80\% & \textbf{29.49} \\
\midrule
\multirow{3}{*}{\textbf{Cross Area}} 
& FG$^2$~\cite{fg2} & \textbf{7.45} & \textbf{4.03} & \textbf{3.33} & \textbf{1.88} & \textbf{30.34}\% & \textbf{81.17}\% & 4.20 \\
& GeoFlow (3-DoF) & 8.53 & 5.68 & 3.87 & 2.75 & 20.00\% & 72.30\% & \textbf{29.49} \\
\bottomrule
\end{tabular}
\end{table*}

\section{Extending GeoFlow to Full 3-DoF}
\label{apdx:3dof_extension}

\textbf{The primary objective of this feasibility study is to demonstrate the inherent modularity and generalizability of the GeoFlow architecture}. We show its capability to seamlessly extend from 2-DoF planar localization (translation-only) to the full 3-DoF pose estimation task (translation $\mathbf{g}_{(x,y)}$ and orientation $\gamma$) with minimal architectural changes and negligible computational overhead. This extension validates that the learned cross-view spatial representations are rich enough to capture both metric displacement and angular alignment between ground and satellite views.

\subsection{Context and Clarification of Main Results}
For fair comparison with existing state-of-the-art methods, the primary 2-DoF evaluation in the main paper (Table~\ref{tab:main_results_kitti}) uses test data where images contain random orientation shifts ($\pm 10^{\circ}$), following other methods~\citep{fg2,CCVPE}. Moreover, the main 2-DoF model in the paper only estimates the translation vector $(x, y)$, without the orientation prediction itself. In this supplementary study, we extend the model to explicitly predict the vehicle's orientation angle $\gamma$, demonstrating GeoFlow's adaptability to the full 3-DoF pose estimation task.

\subsection{Methodology: Architectural Extension}
The extension from 2-DoF to 3-DoF is achieved by appending a lightweight orientation prediction head, $\mathbf{v}_{\gamma}$, to the decoder, thus preserving the core GeoFlow framework.

\subsubsection{Orientation Representation: Unit Circle}
Rather than directly regressing the angle $\gamma$, we adopt a unit circle representation by predicting the pair $(\cos \gamma, \sin \gamma)$. This design choice resolves the inherent periodicity and discontinuity problems associated with angular quantities (e.g., the wrap-around at $\pm 10^{\circ}$). Representing orientation as a point on the unit circle in $\mathbb{R}^2$ ensures continuity, providing smooth and stable gradients for optimization.

The ground-truth orientation $\gamma_{\text{gt}}$ is converted to the unit vector representation \begin{equation}
\mathbf{g}_{\gamma} =
\begin{bmatrix}
\cos(\gamma_{\text{gt}}) \\
\sin(\gamma_{\text{gt}})
\end{bmatrix},
\qquad
\text{where} \quad
\gamma_{\text{gt}} = \gamma_{\text{gt}} \cdot \frac{\pi}{180}.
\label{eq:orientation_unit_vec}
\end{equation}

\subsubsection{Orientation Head and Loss Function}
The orientation prediction head $\mathbf{v}_{\gamma}$ is a simple MLP appended to the decoder: $\mathbf{v}_{\gamma}: \mathbb{R}^{144} \rightarrow \mathbb{R}^{64} \rightarrow \mathbb{R}^{2}$. The output prediction $\mathbf{p}_{\gamma} = [p_{\cos}, p_{\sin}]$ is $\ell_2$-normalized to $\hat{\mathbf{p}}_{\gamma} = \mathbf{p}_{\gamma}/\|\mathbf{p}_{\gamma}\|_2$ to enforce consistency with the unit circle manifold.

We employ a cosine similarity loss $\mathcal{L}_{\gamma}$ to measure the angular distance $\Delta \gamma$ between the predicted $\hat{\mathbf{p}}_{\gamma}$ and ground-truth $\mathbf{g}_{\gamma}$ unit vectors:
\begin{equation}
\mathcal{L}_{\gamma} =
\mathbb{E}\left[
1 -
\frac{
\hat{\mathbf{p}}_{\gamma} \cdot \mathbf{g}_{\gamma}
}{
\left\|\hat{\mathbf{p}}_{\gamma}\right\|
\left\|\mathbf{g}_{\gamma}\right\|
}
\right]
=
\mathbb{E}\left[
1 - \cos(\Delta \gamma)
\right].
\label{eq:orientation_loss}
\end{equation}
This loss is stable ($\mathcal{L}_{\gamma} \in [0, 2]$), differentiable, and correctly handles circular continuity. The final training objective for the 3-DoF model is simply the sum of the original translation losses and the new orientation loss:
\begin{equation}
\mathcal{L}_{\text{3-DoF}}
= \mathcal{L}_{r}
+ \mathcal{L}_{\theta}
+ \mathcal{L}_{\gamma}.
\label{eq:loss_3dof}
\end{equation}

\subsection{Computational Overhead Analysis}
The addition of the orientation head results in a minimal parameter increase, demonstrating high architectural modularity. The base 2-DoF model contains 7.38 M parameters, while the extended 3-DoF model contains 7.39 M parameters, representing an overhead of only $0.13\%$ increase. This minimal overhead is entirely contained within the small new orientation MLP. Furthermore, we measured inference time on a single V100 NVIDIA GPU and found no measurable impact on speed. The base and extended models both maintain an inference rate around $29.49 \text{ FPS}$, confirming that the dominant computation remains in the backbone feature extraction and cross-attention. \textbf{Crucially, this demonstrates that the full 3-DoF capability is achieved with negligible computational overhead, reinforcing the core efficiency argument of the GeoFlow framework.}


\subsection{Quantitative Results and Interpretation}
Quantitative results in Table~\ref{tab:3dof_results} demonstrate that GeoFlow can reliably estimate full 3-DoF pose, achieving comparable accuracy against state-of-the-art method FG2~\citep{fg2} while maintaining a significant advantage in computational efficiency.
The 3-DoF model achieves comparable translation accuracy relative to FG2~\citep{fg2}. For the Same Area, the median localization error is $0.69 \text{m}$, and for the Cross Area, it is $5.68 \text{m}$. This confirms that adding the orientation prediction task does not compromise the learned translation features, highlighting the robustness of the GeoFlow architecture. Moreover, GeoFlow achieves a significant breakthrough in efficiency, processing data at 29.49 FPS. This establishes GeoFlow as the fastest full 3-DoF pose estimation model, operating $7.02\times$ faster than competitive methods like FG2 ($4.20 \text{ FPS}$), a critical advantage for real-time deployment. Furthermore, the orientation prediction results validate the feasibility of extending GeoFlow to estimate heading. On the Same Area, the median orientation error is $1.93^{\circ}$, with $87.8\%$ of predictions within $5^{\circ}$ ($\text{R@5}$ success rate). On the Cross Area, the median error is $2.75^{\circ}$, with a $72.3\%$ $\text{R@5}$ success rate. \textbf{This suggests that GeoFlow can provide fair and practical orientation prediction for real-time applications, especially given the significant speed advantage and minimal architectural overhead. These results confirm GeoFlow's suitability for autonomous navigation tasks requiring accurate and rapid position and heading estimation.}

\begin{table*}[h]
\centering
\caption{Ablation Study: Accuracy vs. Computational Overhead. We report Mean and Median localization error (m) for both Same and Cross-Area splits on KITTI.}
\label{tab:ablation_backbone}
\setlength{\tabcolsep}{4pt}
\begin{tabular}{@{}l c c c c c c c c@{}}
\toprule
& \multicolumn{3}{c}{\textbf{Efficiency}} & \multicolumn{2}{c}{\textbf{Same Area (m)}} & \multicolumn{2}{c}{\textbf{Cross Area (m)}} \\
\cmidrule(lr){2-4} \cmidrule(lr){5-6} \cmidrule(lr){7-8}
Backbone & $\downarrow$Params (M) & $\downarrow$Mem (MiB) & $\uparrow$Speed (FPS) & $\downarrow$Mean & $\downarrow$Median & $\downarrow$Mean & $\downarrow$Median \\
\midrule
EffNet-B0 (Default) & \textbf{7.38} & \textbf{686} & \textbf{29.49} & 0.98 & 0.68 & 8.42 & 5.60 \\
EffNet-B5 (Large) & 54.81 & 1169 & 15.70 & \textbf{0.86} & \textbf{0.60} & \textbf{7.93} & \textbf{5.56} \\
\bottomrule
\end{tabular}
\end{table*}

\subsection{Key Insights and Conclusions}
This feasibility study provides strong evidence for GeoFlow's architectural flexibility. The model was seamlessly extended to a new task (3-DoF pose) with negligible computation overhead. This confirms that GeoFlow provides an efficient and generalizable framework, with learned cross-view representations rich enough to capture both metric displacement and angular alignment, extending to a complete pose 3-DoF estimation pipeline.

\section{Backbone Capacity and Efficiency Trade-Off}
\label{apdx:backbone_ablation}

To address potential concerns regarding model capacity, we conducted an ablation comparing our default lightweight backbone (EfficientNet-B0~\cite{Tan2019EfficientNetRM}) against a larger architecture (EfficientNet-B5~\cite{Tan2019EfficientNetRM}). This study quantifies the fundamental trade-off between absolute accuracy and computational overhead.

The results in \Cref{tab:ablation_backbone} clearly demonstrate the substantial cost required for marginal performance gains. While the larger EfficientNet-B5 backbone is more accurate, providing a slight improvement of $0.08 \text{m}$ in Median Error (Same Area) and $0.04 \text{m}$ gain in Median Error (Cross Area), this comes at a severe computational penalty: the model is $\mathbf{7.4x}$ larger and incurs a $\mathbf{46.7\%}$ reduction in real-time speed. Our GeoFlow is designed to be backbone agnostic and demonstrates clear generalizability, working seamlessly with larger architectures like EfficientNet-B5. However, as mentioned before, this flexibility comes with a significant computational overhead ($\mathbf{7.4x}$ more parameters and $\mathbf{46.7\%}$ slower inference speed), which is not justified by the resulting marginal accuracy gains. This is especially critical for real-world deployment scenarios, as devices like small drones, mobile robots, and embedded systems possess severely limited memory and computational power, meaning they simply cannot run large-scale models like EfficientNet-B5. \textbf{This decisively validates our initial design choice: the lightweight EfficientNet-B0 provides the optimal balance between necessary accuracy and SOTA efficiency, ensuring the model remains viable for real-time deployment.}

\section{Efficiency Analysis and Deployment Considerations}
\label{apdx:efficiency}

This section provides additional analysis of the computational efficiency of GeoFlow and its implications for deployment in real-world localization systems. In practical navigation scenarios, inference speed and memory footprint are critical constraints, particularly for embedded platforms such as drones, mobile robots, and edge vision devices. While improving fine-grained cross-view geolocalization accuracy is important, many real-world applications require models that can operate reliably under strict computational and memory budgets.

\subsection{Architectural Efficiency}

GeoFlow is designed with a strong emphasis on computational efficiency. Compared with existing FG-CVG methods, our model maintains a significantly smaller computational footprint in terms of both FLOPs and memory consumption. This lightweight design enables deployment on resource-constrained hardware while still maintaining competitive localization accuracy.

\subsection{Efficient Iterative Refinement}

The efficiency of the proposed Iterative Refinement Sampling (IRS) algorithm arises from a decoupled computation strategy. The most computationally expensive operations (feature extraction using the EfficientNet backbone and cross-attention-based visual fusion) are performed only once per image pair to obtain the shared visual context representation $\mathbf{f}_{vis}$.

Subsequent refinement iterations operate exclusively on lightweight modules:
\begin{itemize}
    \item Coordinate encoding layers
    \item Small regression MLPs used for flow prediction
\end{itemize}

Since these components represent only a small fraction of the total model computation, multiple refinement iterations can be performed with minimal additional cost. This design allows GeoFlow to benefit from iterative hypothesis refinement without significantly increasing inference latency.

\subsection{Unified Efficiency Benchmark}

Table~\ref{tab:bench} compares GeoFlow with representative FG-CVG approaches in terms of computational cost, memory footprint, and inference latency.

\begin{table}[h]
    \centering
    \footnotesize
    \setlength{\tabcolsep}{3pt}
    \renewcommand{\arraystretch}{0.95}
    \caption{Unified efficiency comparison across representative FG-CVG methods.}
    \label{tab:bench}
    \begin{tabular}{l|cccc}
        \toprule
        \textbf{Model} & \textbf{GFLOPs} & \textbf{VRAM} & \textbf{Inference Time} & \textbf{FPS} \\
        \midrule
        CCVPE & 31.18 & 4730 MiB & 41.7 ms & 24.0 \\
        HC-Net & 11.56 & 1900 MiB & 40.0 ms & 25.0 \\
        \textbf{GeoFlow} & \textbf{7.65} & \textbf{686 MiB} & \textbf{26.0 ms} & \textbf{29.5} \\
        \midrule
        \rowcolor{lightgray} \textit{Gain (vs HC-Net)} & \textbf{-34\%} & \textbf{-64\%} & \textbf{-35\%} & \textbf{+18\%} \\
        \bottomrule
    \end{tabular}
\end{table}

GeoFlow achieves the lowest computational cost among the compared approaches, reducing FLOPs by 34\% and memory consumption by 64\% relative to HC-Net while also achieving lower inference latency. We achieved this stability while still outperforming CCVPE in mean localization accuracy (8.42m vs 9.16m) and maintaining parity with HC-Net. We simply trade peak static precision in rare cases for the operational reliability required to keep the platform airborne. 

\subsection{Deployment Implications}

The reduced computational and memory requirements make GeoFlow particularly suitable for edge deployment scenarios. On embedded platforms such as micro-UAVs or mobile robots, available memory must be shared among multiple perception modules (e.g., obstacle detection, planning, and control). Models with large memory footprints can therefore become impractical even if their theoretical accuracy is high.

By requiring only 686 MiB of VRAM, GeoFlow leaves sufficient memory headroom for other concurrent perception tasks while maintaining video-rate inference (26 ms per frame). Lower computational cost also reduces power consumption and thermal load, which are critical considerations for battery-powered platforms.

Furthermore, lower inference latency directly improves closed-loop navigation stability. For example, for a platform moving at 15 m/s, reducing latency from 60 ms to 26 ms decreases the distance traveled between localization updates by nearly 60\%, improving the responsiveness of downstream navigation systems.

\begin{figure*}[h!]
    \centering
    \includegraphics[width=2.0\columnwidth]{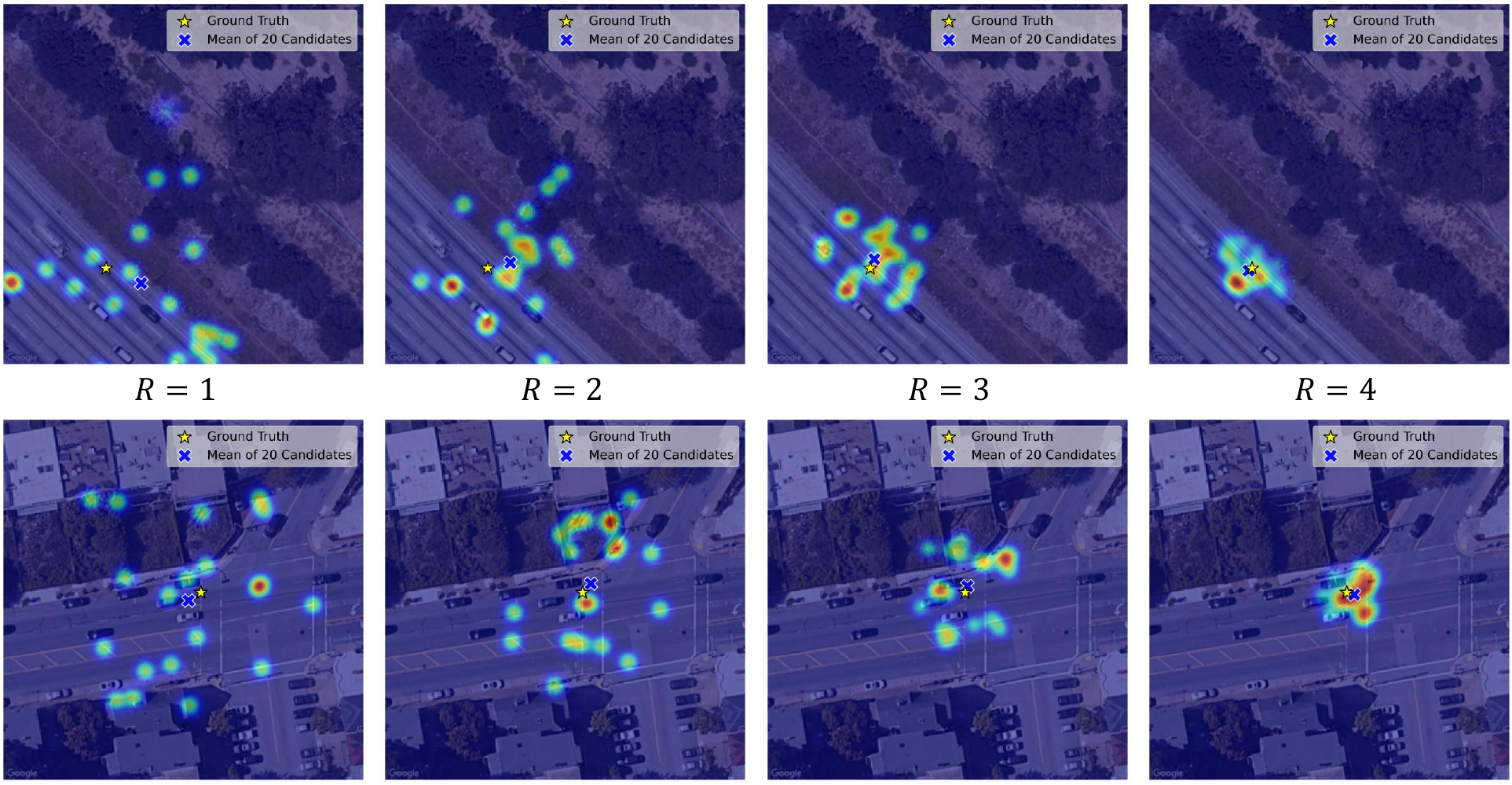} 
    \caption{Additional results for IRS refinement dynamics: Visualization of pose hypotheses distribution over refinement rounds $R \in \{1,2,3,4\}$, demonstrating convergence towards the ground truth and how hypothesis consensus reflects confidence.}
    \label{fig:add_irs_heatmap}
\end{figure*}
\section{Additional Uncertainty Results}
\label{apdx:uncertain}
Figure~\ref{fig:add_irs_heatmap} shows additional results of the uncertainty produced by our \ourmodel{} and how our IRS refines the uncertainty during inference, enhancing accuracy and confidence. These visualizations further support our contribution to the IRS algorithm. \textbf{The convergence of hypotheses from a broad initial distribution to a sharp, final cluster visually proves that IRS is an effective mechanism for building predictive confidence and enhancing overall localization robustness.}

\label{apdx:vis}
\begin{figure*}[t!]
    \centering
    \includegraphics[width=2.0\columnwidth]{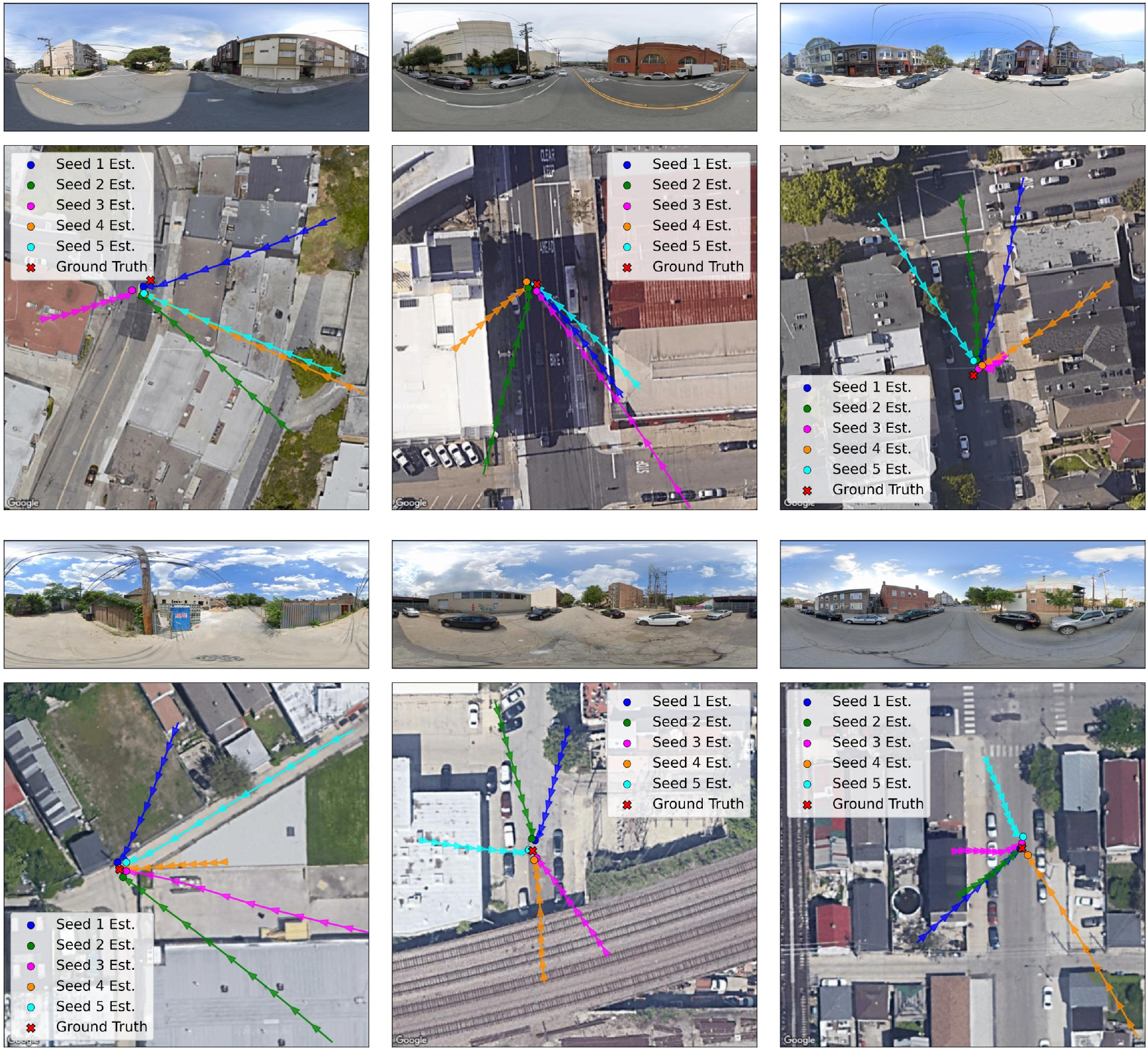} 
    \caption{Additional Qualitative localization examples by \ourmodel{} on the Same-Area VIGOR dataset. The figure visualizes the trajectories generated by the learned regression field by \ourmodel{}: trajectories show how five randomly sampled initial pose hypotheses (seeds) are transformed to their respective predicted locations (colored circular markers), relative to the ground truth (red X).}
    \label{fig:add_qualitative_results}
\end{figure*}

\label{apdx:vis}
\begin{figure*}[t!]
    \centering
    \includegraphics[width=2.0\columnwidth]{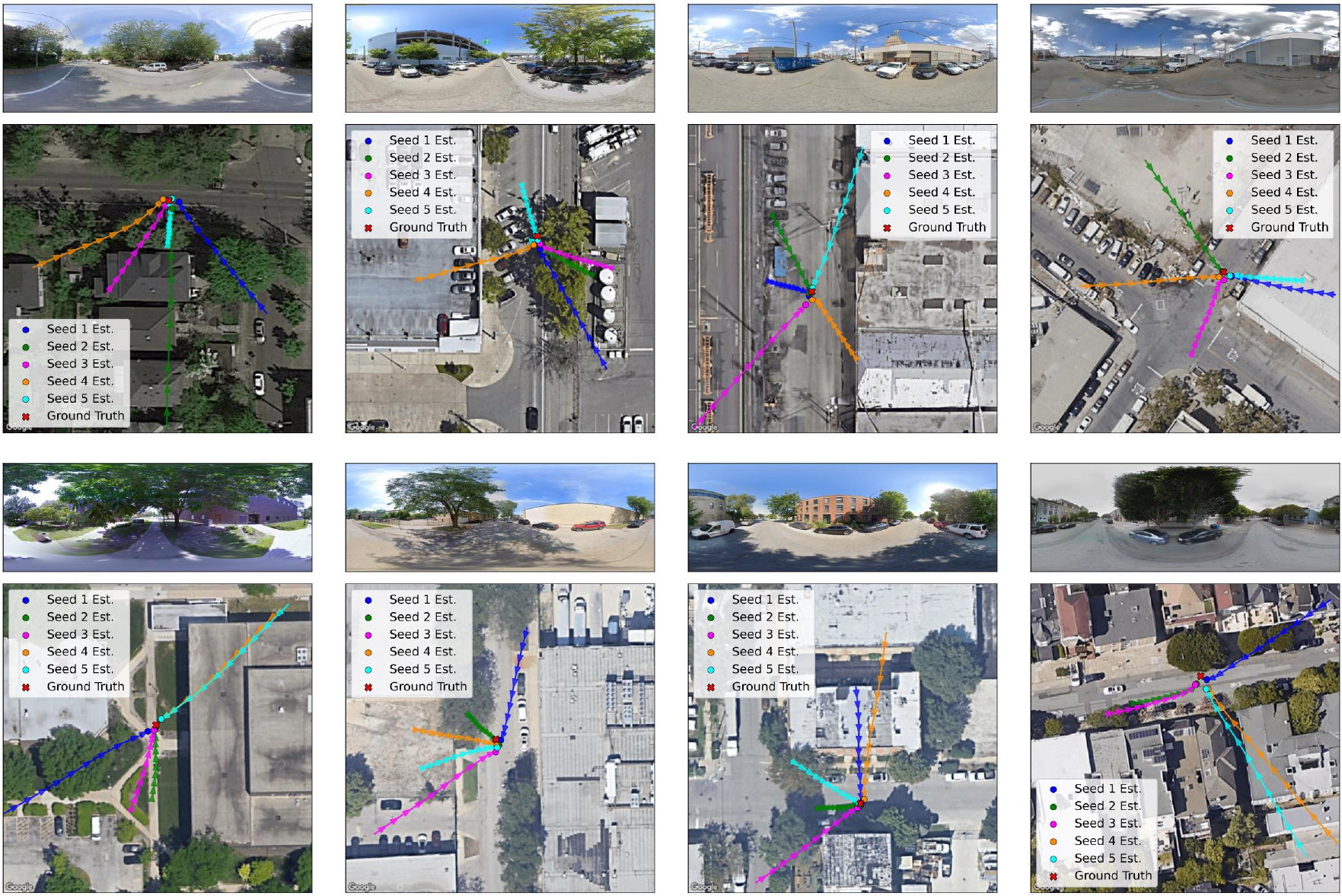} 
    \caption{Additional Qualitative localization examples by \ourmodel{} on the Cross-Area VIGOR dataset. The figure visualizes the trajectories generated by the learned regression field by \ourmodel{}: trajectories show how five randomly sampled initial pose hypotheses (seeds) are transformed to their respective predicted locations (colored circular markers), relative to the ground truth (red X).}
    \label{fig:add_qualitative_results_cross}
\end{figure*}

\section{Additional Qualitative Results}
\label{apdx:vis}

This section provides a deeper qualitative analysis of \ourmodel{}'s performance on the VIGOR dataset through two distinct visual scenarios. Figure~\ref{fig:add_qualitative_results} visualizes the refinement trajectories for the Same-Area setting, demonstrating precise localization within known operational environments. Figure~\ref{fig:add_qualitative_results_cross}, in contrast, showcases the robustness of the learned regression field in the highly challenging Cross-Area setting, highlighting the model's generalization capabilities to unseen areas. \textbf{The combined visual evidence across both known and unseen environments supports the effectiveness of the learned regression field, demonstrating the framework's reliable precision and generalization capabilities.}

\begin{figure*}[t] 
\centering \includegraphics[width=1.7\columnwidth]{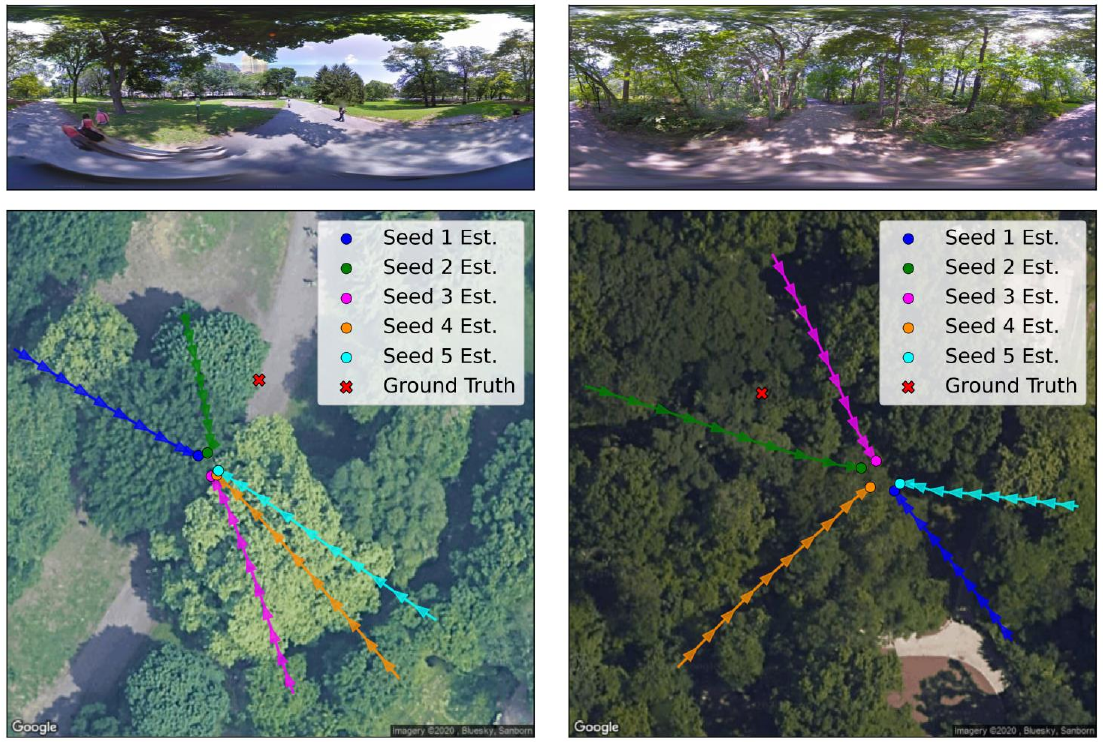}   
\caption{Failure Cases: Localization difficulties in densely vegetated forest environments. The top row shows the ground images, and the bottom row visualizes the refinement trajectories and final converged hypotheses on the corresponding satellite images.}  \label{fig:combined_failure_cases} \end{figure*}

\section{Failure Cases Analysis}
\label{failure cases}

While \ourmodel{} generally demonstrates robust performance, understanding the characteristics of its error distribution is essential. \textbf{This section highlights the specific failure modes responsible for the outliers in our error distribution.}
As illustrated in Figure~\ref{fig:combined_failure_cases}, the majority of these errors are isolated to environments dominated by dense and visually ambiguous vegetation, particularly in the VIGOR Cross-Area setting. The homogeneous textures in the satellite image prevent \ourmodel{} from establishing the necessary feature variance to guide the hypotheses correctly, leading to convergence at an incorrect, distant location.
These catastrophic, isolated failures contribute to the error distribution (i.e., the shift between the Mean and Median error). 

\section{Societal Impact and Limitations}
\label{apdx:impacts}

\noindent\textbf{Societal Impact.} The proposed \ourmodel{} advances the field of autonomous navigation by providing a robust, low-latency alternative to GPS. \textbf{By enabling accurate localization at real-time speeds with minimal computational overhead, our framework is particularly valuable for resource-constrained platforms, such as small delivery drones or mobile search-and-rescue robots operating in GPS-noisy areas, such as metropolitan areas, GPS-denied environments, and mountainous areas.} Furthermore, the robustness gained from iteratively refining diverse poses allows downstream decision-making systems to receive a reliable final location estimate, mitigating risks in critical scenarios like disaster response or urban navigation.

\noindent\textbf{Limitations.} Our method achieves state-of-the-art efficiency, but practical deployment for the FG-CVG task still faces challenges inherent to the domain. 
Performance is sensitive to factors such as the difficulty in accurately predicting location with limited field-of-view (LFOV) ground images, referring to those with a severely restricted viewing angle (unlike the relatively wide LFOV in the KITTI benchmark), where local visual context is severely restricted.

\end{document}